\documentclass[runningheads]{llncs}

\usepackage{eccv}

\usepackage{eccvabbrv}
\usepackage{enumitem}
\usepackage{graphicx}
\usepackage{booktabs}
\usepackage{multirow}
\usepackage[table]{xcolor}
\usepackage{array}
\usepackage{siunitx}
\usepackage{pifont}
\usepackage[accsupp]{axessibility}  

\newcommand{\cmark}{\ding{51}}
\definecolor{firstbg}{RGB}{248,190,190}
\definecolor{secondbg}{RGB}{248,220,182}
\newcommand{\first}[1]{\cellcolor{firstbg}#1}
\newcommand{\second}[1]{\cellcolor{secondbg}#1}
\newcommand{\firsttag}{\colorbox{firstbg}{1st}}
\newcommand{\secondtag}{\colorbox{secondbg}{2nd}}
\usepackage[pagebackref,breaklinks,colorlinks,citecolor=eccvblue,linkcolor=eccvblue]{hyperref}

\usepackage{orcidlink}

\makeatletter
\renewenvironment{abstract}{%
  \list{}{\topsep=2pt\relax\partopsep=0pt\relax\small
  \leftmargin=1cm
  \labelwidth=0pt
  \listparindent=0pt
  \itemindent\listparindent
  \rightmargin\leftmargin}%
  \item[\hskip\labelsep\bfseries\abstractname]}
{\endlist}
\makeatother

\begin{document}

\title{Light ’em Up: Enabling Few-Shot Low-Light 3D Gaussian Splatting with Multi-Scale Explicit Retinex Illumination Decoupling} 


\author{YuHao Yin\inst{1,2,3,4} \and
Zongji Wang\inst{1,2}\thanks{Co-corresponding authors.} \and
Yuanben Zhang\inst{1,2}$^{\star}$ \and
Biqing Li\inst{1,2,3,4} \and
Jiesong Bai\inst{5} \and
Junyi Liu\inst{1,2}}

\authorrunning{Y.~Yin et al.}

\institute{Aerospace Information Research Institute, Chinese Academy of Sciences \and
Key Laboratory of Target Cognition and Application Technology (TCAT) \and
University of Chinese Academy of Sciences \and
School of Electronic, Electrical and Communication Engineering, University of Chinese Academy of Sciences \and
Shanghai Jiao Tong University}

\maketitle

\begin{abstract}
Full 360$^\circ$ novel view synthesis under low-light conditions remains challenging. Insufficient illumination, noise amplification, and view-dependent photometric inconsistencies prevent existing methods from jointly preserving geometric consistency and photorealism. Unsupervised approaches often exhibit color drift under large viewpoint variations, while supervised low-light enhancement models, though effective for 2D tasks, struggle to generalize to new scenes and typically require retraining. To address this issue, we propose MERID-GS, a Multi-Scale Explicit Retinex Illumination-Decoupled Gaussian framework for low-light 360$^\circ$ synthesis. Based on Retinex theory, the method explicitly separates illumination and reflectance, and suppresses noise propagation while enhancing dark-region structures via a learnable gain and Illumination-State-Guided Frequency Gating. Combined with lightweight Reflection Head and 3D Gaussian Splatting, MERID-GS adapts to new scenes with only a few shots and enables stable low-light novel view synthesis from sparse-view observations. In addition, we construct a low-light multi-view dataset covering full 360$^\circ$ scenes for joint evaluation. Thorough experiments across multiple datasets in this area demonstrate that MERID-GS achieves SOTA performance, exhibiting superior cross-scene generalization and view consistency. The source code and pre-trained models are available at \url{https://github.com/YhuoyuH/MERID-GS}.
\end{abstract}

\begin{center}
\includegraphics[width=\linewidth]{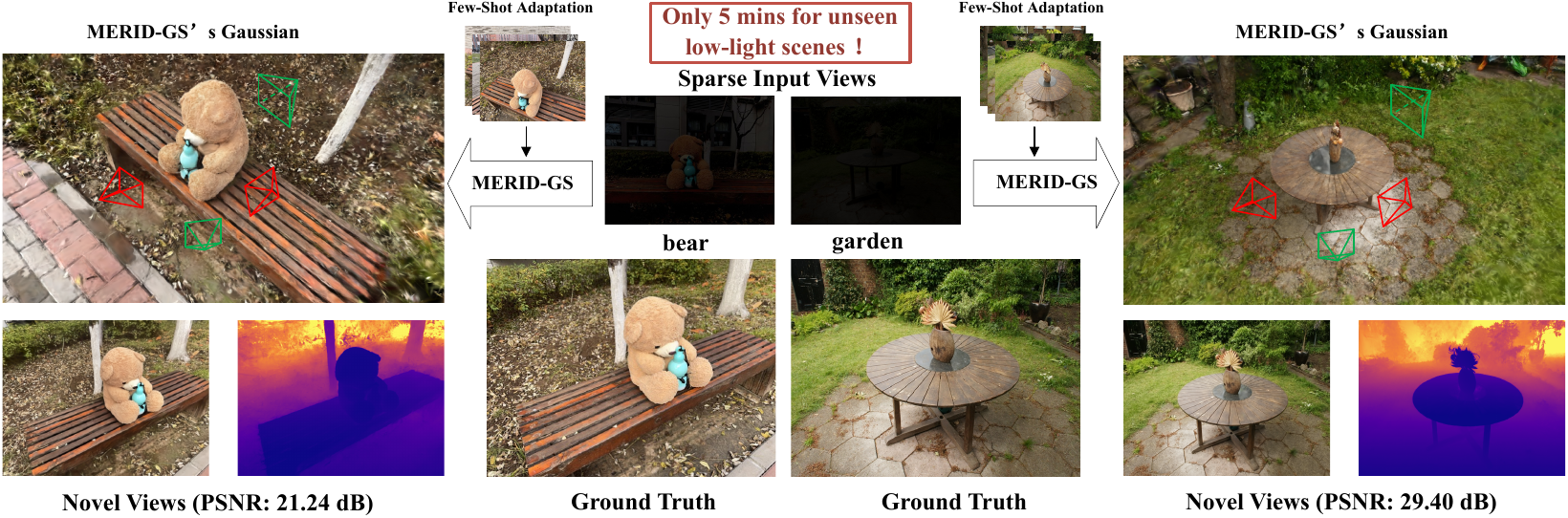}
{\captionsetup{hypcap=false}
\captionof{figure}{MERID-GS lights up new scenes with only 10 normal-light input views for few-shot adaptation and produces high-quality novel view synthesis from sparse-view observations in about 5 minutes end-to-end.}
\label{fig1}}
\end{center}

\section{Introduction}
\label{sec:intro}
Low-light 360$^\circ$ novel view synthesis (NVS) is of significant practical importance in applications such as nighttime autonomous driving, augmented reality, digital twins, and cultural heritage digitization. However, due to the limited dynamic range of camera sensors and complex illumination environments, multi-view images captured under low-exposure conditions often suffer from severe noise, low contrast, and color distortion~\cite{ref1,ref2}. These degradations not only reduce single-view visual quality but also disrupt cross-view appearance consistency, thereby interfering with radiance modeling across viewpoints and degrading the stability and realism of synthesized views.

Recent advances in Neural Radiance Fields (NeRF) and 3D Gaussian Splatting (3DGS)~\cite{ref3,ref4,ref5,ref6,ref7,ref8,ref9,ref10,ref11,ref12,ref13} have significantly improved high-fidelity scene modeling and consistent novel view synthesis. By learning implicit or explicit radiance representations from multi-view images, these methods achieve accurate geometry and appearance reconstruction. However, most existing approaches assume well-exposed and uniformly illuminated inputs. When applied to real-world low-light or exposure-inconsistent scenes, NeRF and 3DGS often misinterpret illumination degradation as radiance or geometric variation, leading to color drift, noise amplification, and unstable view synthesis.

Existing low-light NVS approaches can be broadly categorized into two groups. The first group avoids image preprocessing and instead models illumination degradation during radiance learning. NeRF-W~\cite{ref14} and its extensions~\cite{ref15,ref16,ref17,ref18} introduce view-dependent appearance embeddings to isolate lighting variations. Several low-light NeRF variants~\cite{ref19,ref20,ref2,ref21} further incorporate additional assumptions during volume rendering to restore brightness while maintaining multi-view consistency. However, NeRF-based methods rely on MLP-based implicit representations, resulting in high training and inference costs. In contrast, 3DGS adopts explicit Gaussian ellipsoids for scene representation, offering substantial efficiency advantages~\cite{ref7,ref5,ref22,ref23}. For example, Luminance-GS~\cite{ref24} aligns multi-view illumination via curve adjustment, and LL-Gaussian~\cite{ref25} employs dual-branch Gaussian decomposition for low-light modeling. Nevertheless, whether based on NeRF or 3DGS, such non-preprocessing approaches remain stable under limited viewpoint changes but tend to suffer from severe color distortion and appearance instability when extended to full 360$^\circ$ coverage with sparse inputs.

The second category adopts image preprocessing strategies, performing low-light enhancement prior to NVS. Traditional Retinex-based methods~\cite{ref26} rely on fixed analytical formulations and struggle with real noise, color degradation, and nonlinear camera response. Recent deep learning methods have achieved remarkable progress in 2D low-light enhancement. Unsupervised approaches~\cite{ref38,ref39,ref40,ref41,ref42,ref43} reduce data dependency but often produce unstable enhancement results. Supervised methods~\cite{ref27,ref28,ref29,ref30,ref31,ref32,ref33,ref76,ref72} can achieve better visual quality, but they typically rely on implicit Retinex-style end-to-end decomposition rather than explicitly separating illumination and reflectance. In addition, these methods are usually trained on general-purpose datasets, such as SID~\cite{ref34}, SDSD-indoor~\cite{ref35}, and LOL~\cite{ref36,ref27}. Consequently, they lack scene-specific illumination modeling capability and usually require time-consuming retraining with paired data when applied to unseen scenes, limiting their practicality for cross-scene NVS.

Based on the above observations, we propose \textbf{MERID-GS}, a \textbf{Multi-Scale Explicit Retinex Illumination-Decoupled Gaussian} framework for low-light novel view synthesis. Unlike introducing complex illumination assumptions at the 3D representation stage, we perform explicit illumination modeling in the image domain under the Retinex theory, disentangling illumination and reflectance via a multi-scale decoupling module and constructing a scene-adaptive illumination compensation mechanism. To alleviate noise propagation across views under low-light conditions, we further introduce an Illumination-State-Guided Frequency Gated Attention (IS-FGA) mechanism, and combine it with a lightweight Reflection Head and 3D Gaussian Splatting to achieve stable and consistent 360$^\circ$ novel view synthesis.

Benefiting from the general illumination representation brought by explicit illumination decoupling, as shown in \cref{fig1}, MERID-GS requires only 10 normally lit input views for few-shot adaptation to unseen scenes and achieves high-quality novel view synthesis from sparse-view observations, substantially reducing per-scene training and optimization cost. To systematically evaluate cross-view consistency and generalization, we construct a 360$^\circ$ low-light multi-view dataset, LLD, and conduct experiments on NeRF360~\cite{ref3}, LOM\_full~\cite{ref19}, and LLD. Results on 17 scenes demonstrate that \textbf{MERID-GS} achieves SOTA performance, consistently surpassing existing methods in both synthesis quality and cross-scene generalization.

The main contributions of this work are summarized as follows:
{\setlength{\emergencystretch}{1.5em}
\begin{itemize}[leftmargin=*,topsep=2pt,itemsep=2pt,parsep=0pt,partopsep=0pt]
    \item We propose \textbf{MERID-GS}, a Multi-Scale Explicit Retinex Illumination-Decoupled Gaussian framework for low-light 360$^\circ$ novel view synthesis that enables high-quality rendering on unseen scenes from sparse-view observations.
    
    \item We introduce an Illumination-State-Guided Frequency Gated Attention (IS-FGA) to suppress noise propagation in low-light cross-view modeling and enhance structural representation in dark regions.
    
    \item We construct LLD, a full 360$^\circ$ multi-view low-light dataset for novel view synthesis. Thorough experiments on three datasets demonstrate that our method achieves SOTA performance in synthesis quality and cross-scene generalization.
\end{itemize}
}

\section{Related Work}
\subsection{3D Gaussian Splatting}
The introduction of Neural Radiance Fields (NeRF) has significantly advanced research in novel view synthesis (NVS). NeRF~\cite{ref11} and its extensions~\cite{ref3,ref4,ref44,ref6,ref8,ref45,ref46,ref47} learn implicit radiance representations to generate high-quality views. However, their reliance on dense ray sampling and MLP-based implicit modeling leads to substantial computational overhead, limiting training efficiency and real-time rendering capability. To improve efficiency for NVS, 3D Gaussian Splatting (3DGS)~\cite{ref7} adopts explicit 3D Gaussian ellipsoids and leverages efficient splatting-based rendering, achieving significantly faster training and inference. Building upon this representation, numerous extensions have emerged~\cite{ref5,ref10,ref12,ref48,ref25,ref50,ref51}, including 4DGS~\cite{ref48} for dynamic scenes, 2DGS~\cite{ref49} for enhanced surface modeling, Scaffold-GS~\cite{ref50} for improved rendering efficiency, and Pano2Room~\cite{ref51} for panoramic inputs. Despite their success under well-exposed conditions, these methods generally assume stable illumination. In low-light environments, noise amplification, color distortion, and exposure inconsistency disrupt cross-view radiance consistency, making stable and high-quality 360$^\circ$ novel view synthesis difficult and limiting their applicability in real-world scenarios.

\subsection{Low-light Image Enhancement}
Low-light image enhancement, as a fundamental 2D vision task, has achieved substantial progress in recent years. Early methods, such as histogram equalization~\cite{ref52,ref53,ref54,ref55,ref56} and gamma correction~\cite{ref57,ref58,ref59}, improve brightness by adjusting pixel distributions or contrast without explicitly modeling real illumination, often leading to color distortion and artifacts. To enhance physical interpretability, Retinex-based approaches~\cite{ref60,ref63,ref64,ref61,ref62} decompose images into illumination and reflectance components. However, traditional Retinex methods rely on fixed analytical formulations, making it difficult to model noise and nonlinear camera responses in real low-light images, and frequently amplifying noise during enhancement. With the rise of deep learning, neural networks have significantly improved modeling capacity for low-light enhancement. Unsupervised methods~\cite{ref65,ref66,ref38,ref39,ref40,ref41,ref42,ref43} do not require paired data but often suffer from limited quality and stability due to the lack of effective supervision. In contrast, supervised approaches—including CNN-based models~\cite{ref67,ref68,ref69,ref70,ref71,ref72,ref29,ref73}, vision Transformer-based methods~\cite{ref74,ref75,ref76}, hybrid architectures such as SNR-Net~\cite{ref30}, and recent Mamba-based models~\cite{ref32,ref33}—achieve superior results on benchmark datasets. Nevertheless, these methods typically rely on fixed training datasets and lack scene-specific adaptability. Moreover, most approaches implicitly approximate Retinex through end-to-end networks without explicitly disentangling illumination and reflectance~\cite{ref31,ref32}, limiting cross-scene generalization. As a result, they often require extensive retraining for new scenes, reducing deployment efficiency under data-constrained settings.

\subsection{Low-Light Novel View Synthesis}
Recent low-light novel view synthesis approaches predominantly adopt unsupervised paradigms~\cite{ref19,ref24,ref25}, typically based on NeRF or 3DGS, where illumination degradation is handled directly within the radiance representation stage without image preprocessing. While these methods achieve reasonable performance on datasets with limited viewpoint variation (e.g., LOM\_full), their performance degrades significantly when extended to full 360$^\circ$ coverage. In real-world scenarios, 360$^\circ$ captures are often sparse, further exacerbating instability in cross-view modeling under low-light conditions. These observations suggest that enhancing cross-view appearance consistency via image preprocessing provides a more robust alternative. Our proposed MERID-GS lights up each 360$^\circ$ scene with only about 10 images for few-shot adaptation and enables novel view synthesis from sparse input views, demonstrating strong cross-scene generalization.

\section{Method}
\begin{figure}[t]
\centering
\includegraphics[width=\linewidth]{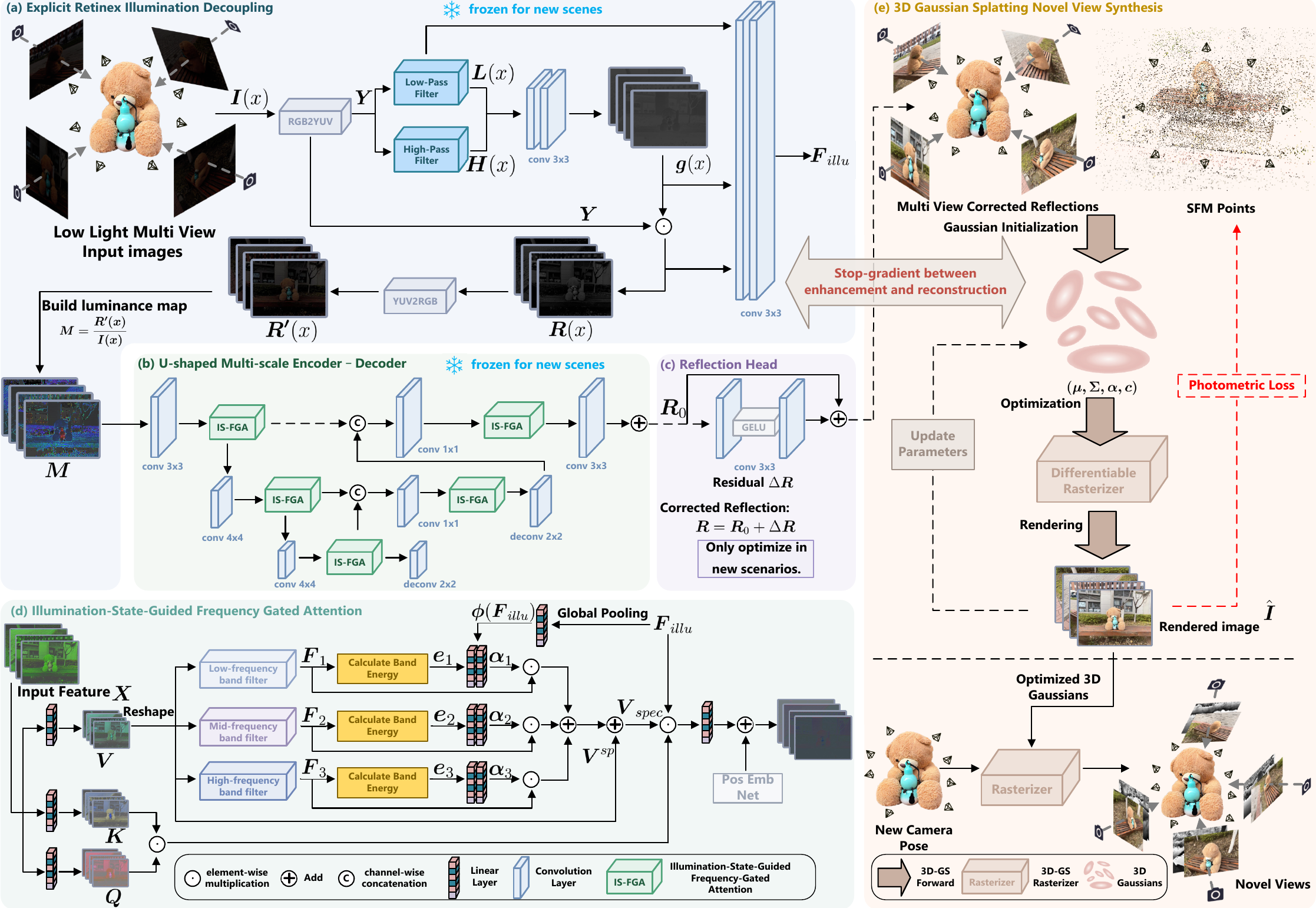}
\caption{Overview of MERID-GS. The first stage (a)(b) corresponds to Multi-scale Explicit Retinex Illumination Decoupling (MERID), which explicitly separates illumination and reflectance based on Retinex theory, and incorporates an Illumination-State-guided Frequency Gated Attention (IS-FGA, see (d)) within a U-Net-like architecture. The second stage (c) is the Reflection Head, which performs lightweight fine-tuning on the reflectance component for unseen scenes. The third stage (e) is a 3DGS-based novel view synthesis module that generates 360$^\circ$ views from the enhanced multi-view images.}
\label{fig2}
\end{figure}
\cref{fig2} illustrates the overall framework of MERID-GS, which consists of three stages. The first stage, \textbf{Multi-scale Explicit Retinex-based Illumination Decoupling (MERID)}, explicitly separates illumination and reflectance and incorporates \textbf{Illumination-State-Guided Frequency Gated Attention (IS-FGA)} to perform denoising and structural enhancement for low-light images. The structures of the Retinex decoupling and IS-FGA modules are shown in \cref{fig2}. 

The second stage is the \textbf{Reflection Head}, which corrects cross-scene color and material deviations and enables few-shot adaptation. The third stage performs novel view synthesis based on \textbf{3D Gaussian Splatting}. Built upon the corrected reflectance representation, it conducts 3D Gaussian modeling to generate high-quality novel views in new scenes.

\subsection{Explicit Retinex Illumination Decoupling}
We propose an explicit illumination–reflectance separation modeling approach based on the Retinex theory. By explicitly decoupling illumination and reflectance in the luminance space, the model learns scene-agnostic illumination compensation strategies, thereby enhancing cross-scene generalization. For unseen scenes, only lightweight adaptation of the reflectance branch is required to recover stable structural representations.

\noindent\textbf{Learnable Regularized Illumination Gain Function.}
To reduce color interference, we first convert the input RGB image into the YUV space and perform illumination modeling only on the luminance channel $Y(x)$, while the chrominance channels are reserved for subsequent color reconstruction.

Conventional Retinex models are formulated in the logarithmic domain, which is prone to numerical instability under low-light conditions. When pixel intensity approaches zero, $\log I(x)$ may lead to gradient explosion. To avoid this issue, we model the reflectance component in the spatial domain:
\begin{equation}
R(x) = \frac{I(x)}{L(x)}.
\end{equation}

An ideal fixed inverse illumination formulation $g(x)=\frac{1}{L(x)}$ is susceptible to noise amplification in real scenarios, ignores camera nonlinearity, and lacks adaptability to varying luminance distributions. To address this limitation, we introduce a learnable regularized gain function:
\begin{equation}
g(x)=g_{\min}+\sigma\big(f_\theta(L(x),H(x))\big)(g_{\max}-g_{\min}),
\end{equation}
where $g_{\min}=0.3$ and $g_{\max}=6$ constrain the dynamic range, and $f_\theta(\cdot)$ denotes a convolutional mapping function. Here, $L(x)$ represents the low-pass luminance component, and $H(x)$ denotes the high-frequency structural response. By jointly modeling luminance and structural information, the design enables adaptive illumination compensation while suppressing noise amplification and preserving reliable structures.

\noindent\textbf{Reflectance Reconstruction and Luminance Mapping.}
The reflectance component is restored in the luminance domain and mapped back to the RGB space:
\begin{equation}
R'(x) = \mathcal{C}^{-1}((Y(x)\cdot g(x)),U,V).
\end{equation}

To explicitly characterize the enhancement strength, we construct a luminance mapping:
\begin{equation}
M = \frac{R'(x)}{I(x)}.
\end{equation}

The resulting map reflects the pixel-wise amplification intensity and serves as an illumination prior to guide subsequent denoising and structural reconstruction modules.

\noindent\textbf{Illumination State Representation.}
To encode the current illumination distribution and enhancement state, we construct an illumination-state feature:
\begin{equation}
\mathbf{F}_{illu}=h_\theta([L,R,\log g]),
\end{equation}
where $h_\theta(\cdot)$ denotes a feature projection network. Introducing $\log g$ compresses the dynamic range, improving training stability and cross-scene generalization.

\subsection{Illumination-State-Guided Frequency Gated Attention}
Initial low-light enhancement often amplifies noise and introduces pseudo structures. 
To improve structural stability, we adopt a U-shaped multi-scale encoder–decoder with attention for denoising and reconstruction. However, in low-light scenes, noise and true structures are highly entangled in the feature space, and global attention aggregation may lead to noise propagation. To address this issue, we propose Illumination-State-Guided Frequency Gated Attention (IS-FGA), which performs dual frequency and spatial filtering prior to attention computation, suppressing noise propagation while enhancing structural representation in dark regions.

\noindent\textbf{Frequency-Conditioned Value Modulation.}
Let the input feature be $\mathbf{X}\in\mathbb{R}^{H\times W\times C}$. In standard multi-head attention, $\mathbf{X}$ is projected into $\mathbf{Q}$, $\mathbf{K}$, and $\mathbf{V}$. We keep $\mathbf{Q}$ and $\mathbf{K}$ unchanged and modulate only the $\mathbf{V}$ branch, thereby constraining the aggregated information without altering the matching relationship. Frequency decomposition is first performed via multiple depthwise separable convolutions:
\begin{equation}
\mathbf{F}_b=\mathcal{H}_b(\mathbf{V}),\quad b=1,\dots,B,
\end{equation}
where $\mathcal{H}_b(\cdot)$ denotes the $b$-th band convolution operator. Convolutions at different scales approximate low-, mid-, and high-frequency responses.

To achieve illumination-aware frequency modulation, we jointly model band energy statistics with the illumination-state condition and directly embed them into the modulation operator. The frequency-modulated value feature is uniformly expressed as:
\begin{equation}
\mathbf{V}_{spec}= \mathbf{V}+ \sum_{b=1}^{B}
f_b\big(\mathbb{E}_{d,H,W}[\mathcal{H}_b(\mathbf{V})], \phi(\mathbf{F}_{illu})\big)
\cdot \mathcal{H}_b(\mathbf{V}),
\end{equation}
where $\mathbb{E}_{d,H,W}$ denotes mean statistics, $\phi(\cdot)$ represents the illumination-state projection function, and $f_b(\cdot)$ is a learnable mapping. This unified formulation enables frequency-selective modulation by jointly considering band energy and illumination conditions, allowing modulation coefficients to adapt dynamically to local illumination distributions. As a result, the model preserves structural components in dark regions while suppressing noise-dominated frequency responses, improving cross-view modeling stability.

\noindent\textbf{Illumination-State-Conditioned Value Gating.}
After frequency modulation, we introduce illumination-state-conditioned gating to perform element-wise modulation and suppress the propagation of noise-dominated regions. Following the dual frequency and spatial filtering, standard attention aggregation is applied:
\begin{equation}
\mathbf{Attention}=\mathrm{Softmax}\left(\frac{\mathbf{K}\mathbf{Q}^\top}{\sqrt{d}}\right)(\mathbf{V}_{spec}\odot \mathbf{F}_{illu}).
\end{equation}

\subsection{Reflection Head for Cross-Scene Appearance Correction}
\begin{figure}[t]
\centering
\includegraphics[width=\linewidth]{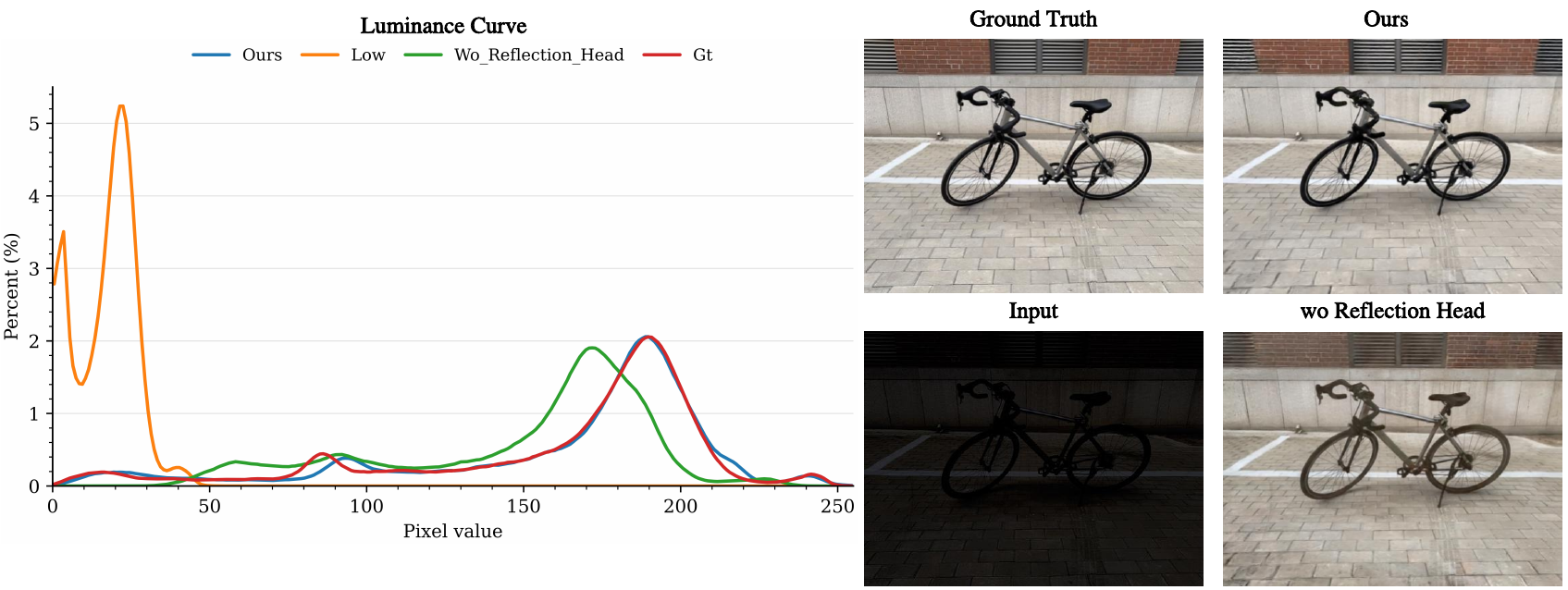}
\caption{Visual comparison on the \emph{bike} scene in LLD, including our result, ground truth , the input, and the model without the Reflection Head, along with the corresponding brightness curves.}
\label{fig3}
\end{figure}
Benefiting from explicit illumination decoupling and the general illumination representation learned from base-scene training, MERID predicts continuous and plausible illumination distributions on unseen scenes, as shown in \cref{fig3}. However, since the material distribution of the target scene is not explicitly modeled, the outputs still exhibit deviations in color consistency and fine details.

To compensate for cross-scene material differences, we introduce a lightweight Reflection Head at the network tail to perform channel-wise nonlinear correction on reflectance features. Considering that real reflectance characteristics are affected by material properties and imaging pipeline nonlinearity, a single linear mapping is insufficient for scene adaptation. We therefore adopt a mapping structure composed of $1\times1$ convolutions and GELU activation to adjust appearance without introducing spatial perturbation.

Let the reflectance feature predicted by MERID be $\mathbf{R}_0 \in \mathbb{R}^{H\times W\times 3}$, the final reflectance output is expressed as:
\begin{equation}
\mathbf{R}=\mathbf{R}_0+\psi(\mathbf{R}_0),
\end{equation}
where $\psi(\cdot)$ denotes a lightweight channel mapping network. Due to its minimal parameter overhead, the module can be adapted with only a few new-scene samples, thereby improving color stability and material consistency under the few-shot setting.

\subsection{3D Gaussian Splatting Novel View Synthesis}
After explicit illumination decoupling and reflection correction, we perform novel view synthesis using 3D Gaussian Splatting (3DGS). Given the reflectance images $\{\mathbf{R}_i\}$ and camera parameters, 3DGS represents the scene with explicit 3D Gaussian ellipsoids and renders images via differentiable splatting. During optimization, both geometric and appearance parameters of the Gaussians are jointly updated, with the loss defined as:
\begin{equation}
\mathcal{L}_{\text{3DGS}} = (1-\lambda)\mathcal{L}_1 + \lambda \mathcal{L}_{\text{D-SSIM}}.
\label{eq:3dgs_loss}
\end{equation}
Benefiting from the stabilized reflectance representation, the model converges rapidly during the few-shot adaptation stage, enabling illumination adaptation to new scenes and performing novel view synthesis under sparse-view observations.

\section{Experiments}
\label{sec:blind}

\subsection{Datasets and Experimental Details}
\noindent\textbf{Implementation Details.}
This section introduces the LLD dataset and the experimental results of \textbf{MERID-GS}. The model is implemented with PyTorch~\cite{ref78} and GS-Splat~\cite{ref37}, and trained and evaluated on an NVIDIA RTX 4090 (24GB). Experiments are conducted on NeRF360, LOM\_full, and LLD. Detailed training configurations are provided in the supplementary material. For fair comparison, all baseline methods adopt their official implementations and recommended settings during base training. In the new-scene fine-tuning stage, all methods are equipped with the same Reflection Head architecture and evaluated under identical training strategies and hyperparameter configurations.

\noindent\textbf{Dataset.}
The existing low-light dataset LOM\_full provides limited viewpoints, which are insufficient for low-light preprocessing and multi-view modeling. To address this limitation, we construct a new low-light multi-view dataset, termed LLD. LLD contains six scenes, five for evaluation and one \textit{bench} scene for base training, with 136--213 multi-view images per scene. We allocate 75\% of the images for preprocessing training and the remaining 25\% for novel view reconstruction, among which 12.5\% are used for quantitative testing. Training and evaluation subsets are divided by uniformly spaced viewpoints to ensure balanced distribution. 

For unseen scenes, 10 views are uniformly sampled from the 75\% preprocessing-training split for few-shot adaptation of the Reflection Head, while the 3D Gaussian Splatting reconstruction is conducted using the reconstruction subset (25\% of the images) following the dataset split protocol.

Data acquisition covers the full 360$^\circ$ range. For each viewpoint, paired low-light and normal-light images are synchronously captured using the \textit{ProCamera} bracketing function, with a tripod to maintain consistent poses. The original resolution is $4032\times3024$, and all images are downsampled to $504\times378$ during training. Camera poses are estimated in \textit{COLMAP} using normal-light images. In addition to low-light and normal-light pairs, LLD also provides overexposed images to support research under complex illumination conditions. Furthermore, we conduct evaluations on NeRF360 and LOM\_full, and generate corresponding low-light versions for NeRF360. The data splits of all three datasets are unified to ensure fair comparison.

\begin{table*}[t]
\centering
\caption{Quantitative comparisons on the NeRF360, LOM\_full, and LLD datasets. All methods are evaluated without the Reflection Head. The best and second-best results are denoted by \firsttag and \secondtag, respectively.}
\label{tab:1}
\scriptsize
\setlength{\tabcolsep}{3pt}
\renewcommand{\arraystretch}{1.15}
\resizebox{\textwidth}{!}{%
\begin{tabular}{l|ccc|ccc|ccc|ccc}
\toprule
\multirow{2}{*}{Methods}
& \multicolumn{3}{c|}{NeRF360}
& \multicolumn{3}{c|}{LOM\_full}
& \multicolumn{3}{c|}{LLD}
& \multicolumn{3}{c}{Average} \\
\cmidrule(lr){2-4}\cmidrule(lr){5-7}\cmidrule(lr){8-10}\cmidrule(lr){11-13}
& PSNR & SSIM & LPIPS
& PSNR & SSIM & LPIPS
& PSNR & SSIM & LPIPS
& PSNR & SSIM & LPIPS \\
\midrule
\multicolumn{13}{c}{\textbf{Unsupervised Methods}} \\
\midrule
Z-DCE~\cite{ref38} + GS (CVPR 2020)      & 13.10 & 0.423 & 0.293 & 13.60 & 0.716 & 0.267 & 10.25 & 0.473 & 0.346 & 12.32 & 0.537 & 0.302 \\
GS + Z-DCE~\cite{ref38}                  & 12.63 & 0.331 & 0.385 & 13.18 & 0.595 & 0.457 & 10.18 & 0.348 & 0.491 & 11.99 & 0.425 & 0.444 \\
LLVE~\cite{ref40} + GS (CVPR 2021)       & 16.98 & 0.516 & 0.463 & 16.31 & 0.725 & 0.424 & 14.08 & 0.583 & 0.372 & 15.79 & 0.608 & 0.420 \\
GS + LLVE~\cite{ref40}                   & 16.66 & 0.487 & 0.519 & 16.17 & 0.657 & 0.528 & 13.59 & 0.435 & 0.537 & 15.47 & 0.526 & 0.528 \\
SCI~\cite{ref39} + GS (CVPR 2022)        & 11.76 & 0.308 & 0.358 & 13.09 & 0.640 & 0.298 & 10.71 & 0.453 & 0.352 & 11.85 & 0.467 & 0.336 \\
GS + SCI~\cite{ref39}                    & 11.41 & 0.248 & 0.436 & 12.39 & 0.527 & 0.478 & 10.34 & 0.335 & 0.490 & 11.38 & 0.370 & 0.468 \\
SGZ~\cite{ref41} + GS (WACV 2022)        & 13.56 & 0.449 & 0.279 & 14.41 & 0.737 & 0.284 & 11.50 & 0.522 & 0.328 & 13.16 & 0.569 & 0.297 \\
GS + SGZ~\cite{ref41}                    & 12.56 & 0.237 & 0.432 & 13.33 & 0.524 & 0.472 & 10.32 & 0.193 & 0.510 & 12.07 & 0.318 & 0.471 \\
NeRCo~\cite{ref42} + GS (ICCV 2023)      & 18.37 & 0.528 & 0.324 & 18.58 & 0.777 & 0.312 & 13.98 & 0.477 & 0.351 & 16.98 & 0.594 & 0.329 \\
GS + NeRCo~\cite{ref42}                  & 18.24 & 0.507 & 0.393 & 18.44 & 0.668 & 0.420 & 14.03 & 0.444 & 0.424 & 16.90 & 0.540 & 0.412 \\
3DGS~\cite{ref7} (SIGGRAPH 2023)        &  8.89 & 0.057 & 0.742 &  6.89 & 0.151 & 0.714 &  5.97 & 0.104 & 0.583 &  7.25 & 0.104 & 0.680 \\
QuadPrior~\cite{ref43} + GS (CVPR 2024)  & \second{19.89} & \second{0.669} & \second{0.231} & \second{20.19} & \second{0.822} & \second{0.263} & 12.77 & 0.574 & 0.334 & \second{17.62} & \second{0.688} & \second{0.276} \\
GS + QuadPrior~\cite{ref43}              & 18.14 & 0.555 & 0.342 & 19.48 & 0.702 & 0.436 & 12.91 & 0.440 & 0.465 & 16.84 & 0.566 & 0.414 \\
Luminance-GS~\cite{ref24} (CVPR 2025)    & 11.72 & 0.484 & 0.397 & 18.64 & 0.805 & 0.283 & 16.29 & 0.611 & 0.331 & 15.55 & 0.633 & 0.337 \\
\midrule
\multicolumn{13}{c}{\textbf{Supervised Methods}} \\
\midrule
MIRNet~\cite{ref72} + GS (ECCV 2020)          & 16.43 & 0.555 & 0.396 & 17.29 & 0.770 & 0.350 & 17.98 & 0.668 & 0.315 & 17.23 & 0.664 & 0.354 \\
GS + MIRNet~\cite{ref72}                      & 17.01 & 0.519 & 0.468 & 16.90 & 0.678 & 0.493 & 17.04 & 0.502 & 0.447 & 16.98 & 0.566 & 0.469 \\
Restormer~\cite{ref76} + GS (CVPR 2022)       & 16.13 & 0.568 & 0.343 & 16.81 & 0.767 & 0.357 & 18.92 & 0.709 & 0.252 & 17.29 & 0.681 & 0.317 \\
GS + Restormer~\cite{ref76}                   & 17.14 & 0.517 & 0.443 & 17.20 & 0.688 & 0.479 & 17.23 & 0.506 & 0.434 & 17.19 & 0.570 & 0.452 \\
SNR-Net~\cite{ref30} + GS (CVPR 2022)         & 13.45 & 0.478 & 0.414 & 16.50 & 0.761 & 0.339 & 19.31 & 0.631 & 0.303 & 16.42 & 0.623 & 0.352 \\
GS + SNR-Net~\cite{ref30}                     & 13.69 & 0.433 & 0.460 & 17.21 & 0.583 & 0.540 & 17.73 & 0.439 & 0.461 & 16.21 & 0.485 & 0.487 \\
Retinexformer~\cite{ref31} + GS (ICCV 2023)   & 16.11 & 0.549 & 0.368 & 16.21 & 0.762 & 0.370 & 19.09 & \second{0.711} & 0.251 & 17.14 & 0.674 & 0.330 \\
GS + Retinexformer~\cite{ref31}               & 16.97 & 0.509 & 0.436 & 16.46 & 0.673 & 0.490 & 17.22 & 0.503 & 0.439 & 16.88 & 0.562 & 0.455 \\
MambaLLIE~\cite{ref32} + GS (NeurIPS 2024)    & 15.57 & 0.547 & 0.377 & 17.17 & 0.745 & 0.381 & 19.09 & \second{0.711} & \second{0.246} & 17.28 & 0.668 & 0.335 \\
GS + MambaLLIE~\cite{ref32}                   & 16.35 & 0.498 & 0.445 & 16.85 & 0.643 & 0.493 & 17.32 & 0.502 & 0.442 & 16.84 & 0.548 & 0.460 \\
Wave-Mamba~\cite{ref33} + GS (MM 2024)        & 15.12 & 0.571 & 0.347 & 14.08 & 0.743 & 0.373 & \second{19.66} & 0.705 & 0.252 & 16.29 & 0.673 & 0.324 \\
GS + Wave-Mamba~\cite{ref33}                  & 15.49 & 0.510 & 0.422 & 14.70 & 0.643 & 0.470 & 17.47 & 0.501 & 0.431 & 15.89 & 0.551 & 0.441 \\
LFPVS~\cite{ref77} + GS (ICCV 2025)           & 14.51 & 0.497 & 0.466 & 14.26 & 0.725 & 0.411 & 19.10 & 0.670 & 0.316 & 15.96 & 0.631 & 0.398 \\
GS + LFPVS~\cite{ref77}                       & 15.15 & 0.469 & 0.521 & 14.74 & 0.647 & 0.517 & 17.62 & 0.494 & 0.503 & 15.84 & 0.537 & 0.514 \\
\midrule
\textbf{Ours}                    & \first{24.05} & \first{0.753} & \first{0.146} & \first{29.60} & \first{0.891} & \first{0.160} & \first{21.43} & \first{0.743} & \first{0.191} & \first{25.03} & \first{0.796} & \first{0.166} \\
\bottomrule
\end{tabular}
}
\end{table*}

\begin{table*}[t]
\centering
\caption{Quantitative comparisons of supervised methods on the NeRF360, LOM\_full, and LLD datasets. For fair comparison, all supervised methods are equipped with the Reflection Head.}
\label{tab:2}
\scriptsize
\setlength{\tabcolsep}{3pt}
\renewcommand{\arraystretch}{1.15}
\resizebox{\textwidth}{!}{%
\begin{tabular}{l|ccc|ccc|ccc|ccc}
\toprule
\multirow{2}{*}{Methods}
& \multicolumn{3}{c|}{NeRF360}
& \multicolumn{3}{c|}{LOM\_full}
& \multicolumn{3}{c|}{LLD}
& \multicolumn{3}{c}{Average} \\
\cmidrule(lr){2-4}\cmidrule(lr){5-7}\cmidrule(lr){8-10}\cmidrule(lr){11-13}
& PSNR & SSIM & LPIPS
& PSNR & SSIM & LPIPS
& PSNR & SSIM & LPIPS
& PSNR & SSIM & LPIPS \\
\midrule
MIRNet~\cite{ref72} + GS (ECCV 2020)          & 20.33 & 0.636 & 0.359 & 22.01 & 0.822 & 0.317 & 20.20 & 0.711 & 0.264 & 20.85 & 0.723 & 0.313 \\
GS + MIRNet~\cite{ref72}                      & 19.40 & 0.528 & 0.449 & 20.50 & 0.709 & 0.453 & 18.01 & 0.503 & 0.451 & 19.30 & 0.580 & 0.451 \\
Restormer~\cite{ref76} + GS (CVPR 2022)       & 19.88 & 0.660 & 0.290 & 22.93 & 0.834 & 0.265 & 20.36 & 0.726 & 0.234 & 21.06 & 0.740 & 0.263 \\
GS + Restormer~\cite{ref76}                   & 19.29 & 0.545 & 0.383 & 20.66 & 0.706 & 0.408 & 18.21 & 0.512 & 0.430 & 19.39 & 0.588 & 0.407 \\
SNR-Net~\cite{ref30} + GS (CVPR 2022)         & 18.64 & 0.560 & 0.372 & 23.32 & 0.814 & 0.260 & 19.29 & 0.634 & 0.306 & 20.42 & 0.669 & 0.313 \\
GS + SNR-Net~\cite{ref30}                     & 18.40 & 0.489 & 0.428 & 19.58 & 0.588 & 0.508 & 17.87 & 0.445 & 0.461 & 18.62 & 0.507 & 0.466 \\
Retinexformer~\cite{ref31} + GS (ICCV 2023)   & \second{23.71} & \second{0.743} & \second{0.162} & \second{29.04} & \second{0.882} & \second{0.191} & \second{20.97} & \second{0.732} & \second{0.216} & \second{24.57} & \second{0.786} & \second{0.190} \\
GS + Retinexformer~\cite{ref31}               & 21.63 & 0.617 & 0.283 & 22.74 & 0.746 & 0.394 & 18.22 & 0.512 & 0.422 & 20.86 & 0.625 & 0.366 \\
MambaLLIE~\cite{ref32} + GS (NeurIPS 2024)    & 19.65 & 0.637 & 0.318 & 21.79 & 0.812 & 0.291 & 19.82 & 0.713 & 0.250 & 20.42 & 0.721 & 0.286 \\
GS + MambaLLIE~\cite{ref32}                   & 19.08 & 0.528 & 0.409 & 20.03 & 0.679 & 0.433 & 17.96 & 0.507 & 0.441 & 19.02 & 0.571 & 0.428 \\
Wave-Mamba~\cite{ref33} + GS (MM 2024)        & 15.10 & 0.567 & 0.355 & 14.16 & 0.749 & 0.355 & 19.59 & 0.705 & 0.251 & 16.28 & 0.674 & 0.320 \\
GS + Wave-Mamba~\cite{ref33}                  & 15.49 & 0.510 & 0.422 & 14.70 & 0.643 & 0.470 & 17.47 & 0.501 & 0.431 & 15.89 & 0.551 & 0.441 \\
LFPVS~\cite{ref77} + GS (ICCV 2025)           & 20.11 & 0.600 & 0.400 & 22.66 & 0.825 & 0.304 & 20.29 & 0.686 & 0.301 & 21.02 & 0.704 & 0.335 \\
GS + LFPVS~\cite{ref77}                       & 19.33 & 0.514 & 0.475 & 19.71 & 0.691 & 0.465 & 18.12 & 0.498 & 0.502 & 19.05 & 0.568 & 0.481 \\
\midrule
\textbf{Ours}                    & \first{24.05} & \first{0.753} & \first{0.146} & \first{29.60} & \first{0.891} & \first{0.160} & \first{21.43} & \first{0.743} & \first{0.191} & \first{25.03} & \first{0.796} & \first{0.166} \\
\bottomrule
\end{tabular}
}
\end{table*}

\subsection{Comparison with Previous Methods}
\noindent\textbf{Baselines.}
For comparison, we adopt Luminance-GS~\cite{ref24} and the original 3DGS as representative unsupervised methods for low-light novel view synthesis. We further include several supervised~\cite{ref30,ref31,ref32,ref33,ref76,ref72,ref77} and unsupervised~\cite{ref38,ref39,ref40,ref41,ref42,ref43} image preprocessing-based low-light enhancement methods, which are combined with 3DGS for evaluation. For preprocessing approaches, two configurations are considered: (1)feeding enhanced images into 3DGS (+3DGS), and (2)applying enhancement as post-processing on 3DGS outputs (3DGS+), to analyze the impact of different low-light processing strategies on synthesis performance.

To ensure fair comparison, we design two evaluation protocols. In the first setting, supervised methods retain original architectures and perform zero-shot inference on new scenes (\cref{tab:1}). In the second setting, to eliminate input discrepancies, we equip all supervised methods with the same Reflection Head and conduct fine-tuning and evaluation under unified architectures and training strategies (\cref{tab:2}). Results on the three datasets are presented in the main paper, with detailed per-scene metrics in the supplementary material.

\noindent\textbf{Quantitative Comparison.}
We present quantitative comparisons between our method and multiple SOTA low-light methods in \cref{tab:1} and \cref{tab:2}. The evaluation is conducted on NeRF360, LOM\_full, and LLD using PSNR, SSIM, and LPIPS. \cref{tab:1} reports results under the zero-shot setting, where all supervised baselines are trained on the base scene (\textit{bench}) with official configurations and directly evaluated on unseen scenes. Both 3DGS-based unsupervised methods and pipelines combining 2D low-light enhancement models with 3DGS exhibit limited cross-scene generalization, with PSNR values mostly ranging from 15–18 dB. In contrast, MERID-GS achieves an average PSNR of \textbf{25.03} dB across the three datasets, significantly outperforming all baselines. It also attains the best performance on SSIM (\textbf{0.796}) and LPIPS (\textbf{0.166}).

To ensure consistent input conditions, \cref{tab:2} equips all supervised methods with the same Reflection Head and fine-tunes them under identical training strategies. Although most methods show improvements, MERID-GS achieves the best performance across all metrics on all datasets. Compared with the strongest supervised baseline, Retinexformer~\cite{ref31} + GS, it still improves the average PSNR by approximately \textbf{0.5} dB while maintaining the best SSIM and LPIPS. These results demonstrate that MERID-GS achieves stable and high-quality low-light novel view synthesis with only a few input views, outperforming existing methods in structural consistency, perceptual quality, and cross-scene generalization.

\begin{figure}[t]
\centering
\includegraphics[width=\linewidth]{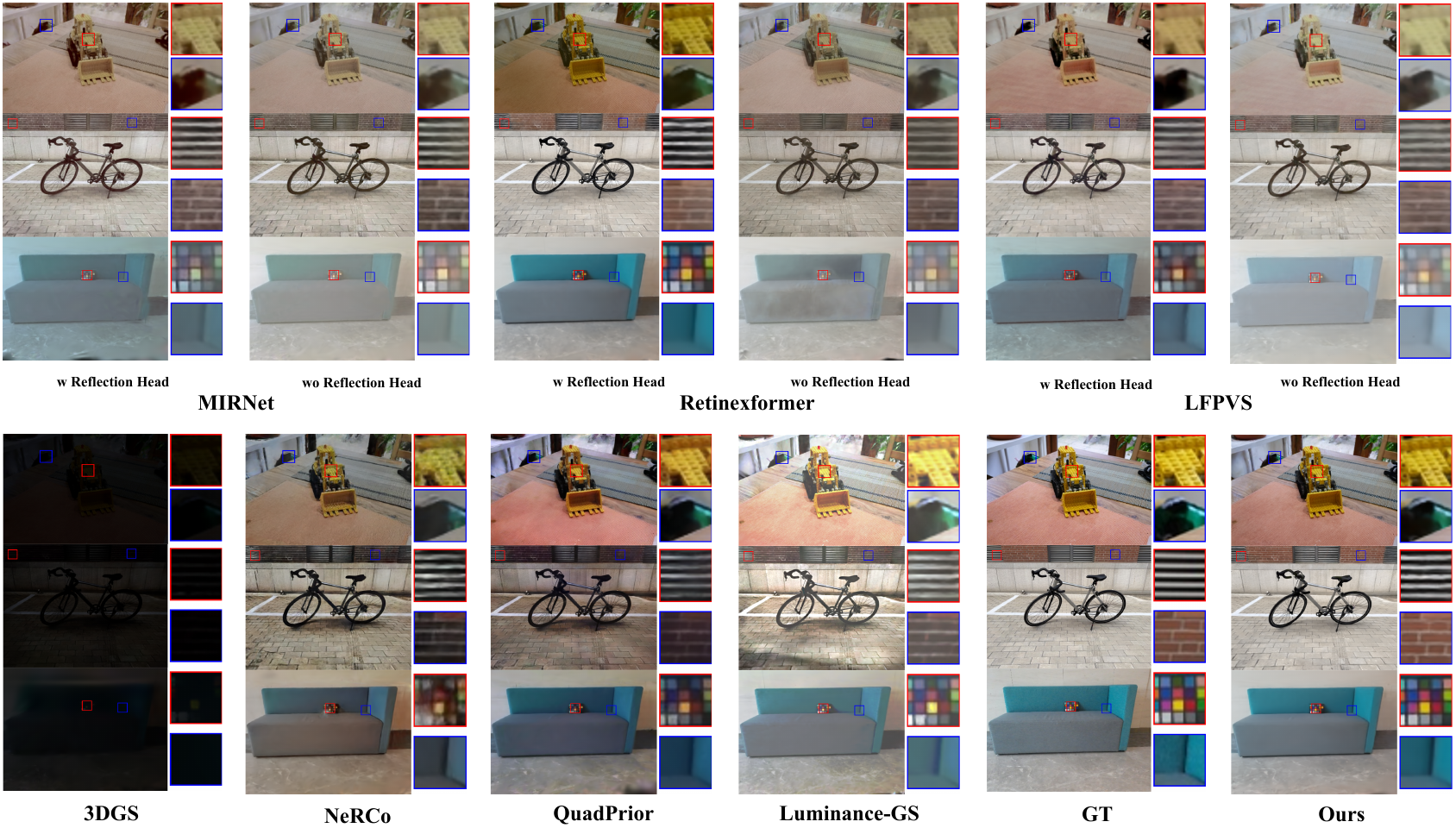}
\caption{Qualitative comparisons on representative scenes (bike, sofa, kitchen) from the NeRF360, LOM\_full, and LLD datasets, against representative baseline methods.}
\label{fig4}
\end{figure}
\begin{figure}[t]
  \centering
  \includegraphics[width=\linewidth]{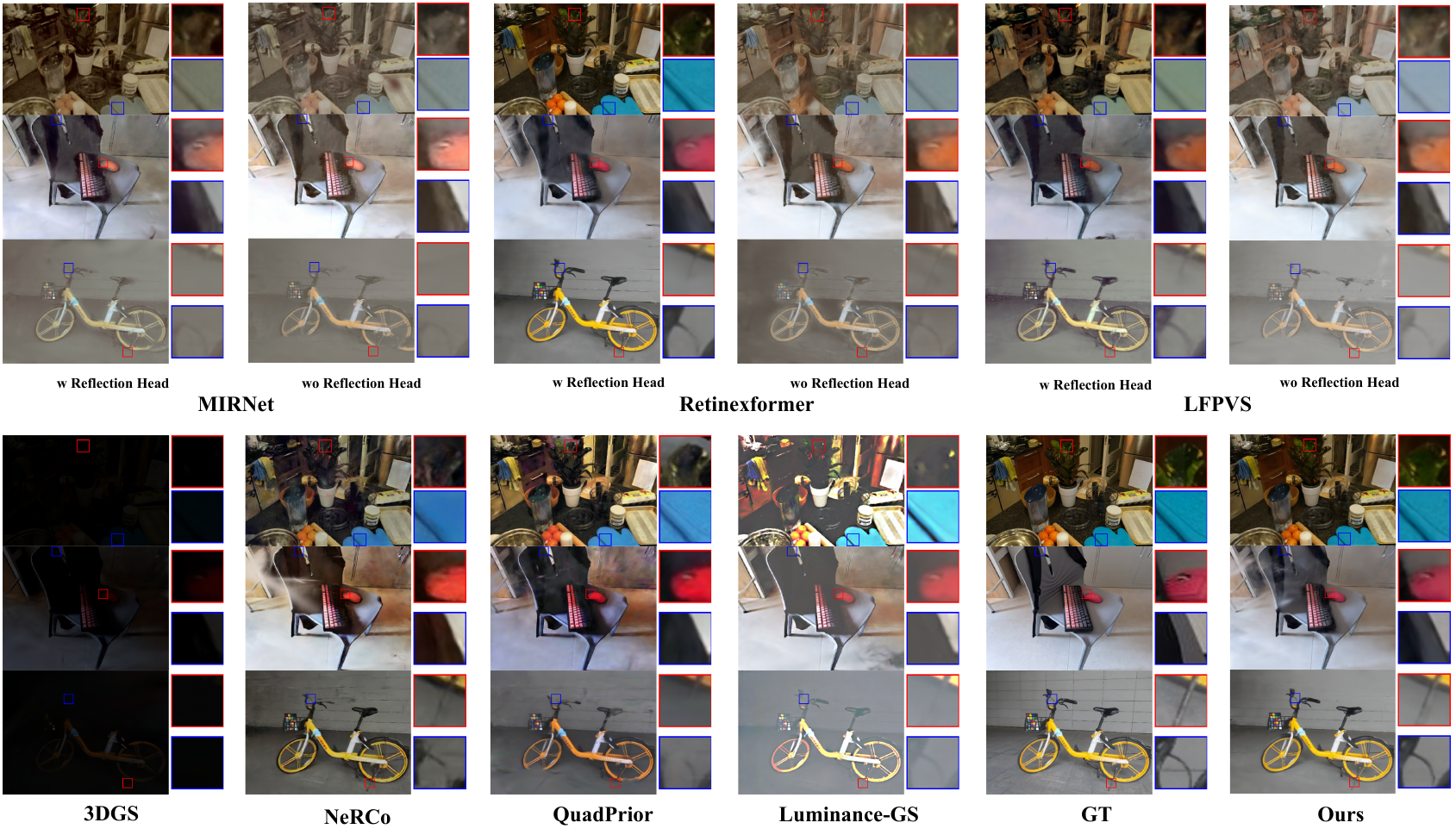}
  \caption{Qualitative comparisons on representative scenes (bike, chair, and counter) from the NeRF360, LOM\_full, and LLD datasets, against representative baseline methods.}
  \label{fig5}
\end{figure}

\noindent\textbf{Qualitative Comparison.}
For qualitative evaluation, \cref{fig4,fig5} show visual comparisons with representative unsupervised and supervised methods. Since the 3DGS+ post-processing scheme performs notably worse than the +3DGS preprocessing setting, we only compare with supervised +3DGS approaches. As shown in the figures, unsupervised methods and supervised models without the Reflection Head suffer from severe color distortion and unstable appearance in unseen scenes. Even supervised methods with a Reflection Head, such as MIRNet~\cite{ref72}, MambaLLIE~\cite{ref32}, and LFPVS~\cite{ref77}, remain prone to color shifts, noise amplification, and artifacts. Although Retinexformer~\cite{ref31} achieves PSNR comparable to ours, its implicit illumination decoupling limits cross-scene compensation, leading to weaker detail recovery and structural consistency. In contrast, MERID-GS generalizes more robustly to unseen scenes, improving brightness while preserving color consistency and geometric structure, and achieves superior visual quality across multiple scenes.

\subsection{Ablation Study}
\begin{table*}[t]
\centering
\footnotesize
\renewcommand{\arraystretch}{1.15}
\setlength{\tabcolsep}{3pt}
\begin{minipage}[t]{0.49\textwidth}
\centering
\caption{Module-level ablation results on the NeRF360, LOM\_full, and LLD datasets, evaluating the contribution of each component across different datasets.}
\label{tab:3}
\resizebox{\linewidth}{!}{%
\begin{tabular}{
  lcccc|
  S[table-format=2.2] S[table-format=1.3] S[table-format=1.3] S[table-format=3.1]
}
\toprule
\textbf{Experiment} &
\textbf{Baseline} &
\textbf{ERID} &
\textbf{IS-FGA} &
\textbf{RF Head} &
\textbf{PSNR} & \textbf{SSIM} & \textbf{LPIPS} & \textbf{TIME(s)} \\
\midrule

\textit{Setting1} & \cmark & & & &
17.12 & 0.671 & 0.317 & 89.1 \\

\textit{Setting2} & \cmark & \cmark & & &
17.13 & 0.676 & 0.314 & 88.7 \\

\textit{Setting3} & \cmark & \cmark & & \cmark &
24.48 & 0.793 & 0.171 & 240.9 \\

\textit{Setting4} & \cmark & \cmark & \cmark & &
17.56 & 0.683 & 0.303 & 97.7 \\

\textbf{Ours} & \cmark & \cmark & \cmark & \cmark &
\multicolumn{1}{c}{\first{25.03}} & \multicolumn{1}{c}{\first{0.796}} & \multicolumn{1}{c}{\first{0.166}} & 268.3 \\

\bottomrule
\end{tabular}
}
\end{minipage}\hfill
\begin{minipage}[t]{0.49\textwidth}
\centering
\caption{Ablation results of ERID and IS-FGA on the LLD base scene (bench). The Reflection Head is not included, as it is not used during the base training stage.}
\label{tab:4}
\resizebox{\linewidth}{!}{%
\begin{tabular}{
  lccc
  S[table-format=2.2] S[table-format=1.3] S[table-format=1.3]
}
\toprule
\textbf{Experiment} &
\textbf{Baseline} &
\textbf{ERID} &
\textbf{IS-FGA} &
\textbf{PSNR} & \textbf{SSIM} & \textbf{LPIPS} \\
\midrule

\textit{Setting1} & \cmark & & &
22.45 & 0.752 & 0.139 \\

\textit{Setting2} & \cmark & \cmark & &
22.54 & 0.753 & 0.139 \\

\textbf{Ours} & \cmark & \cmark & \cmark &
\multicolumn{1}{c}{\first{22.73}} & \multicolumn{1}{c}{\first{0.757}} & \multicolumn{1}{c}{\first{0.134}} \\

\bottomrule
\end{tabular}
}
\end{minipage}
\end{table*}

\begin{figure}[t]
  \centering
  \includegraphics[width=\linewidth]{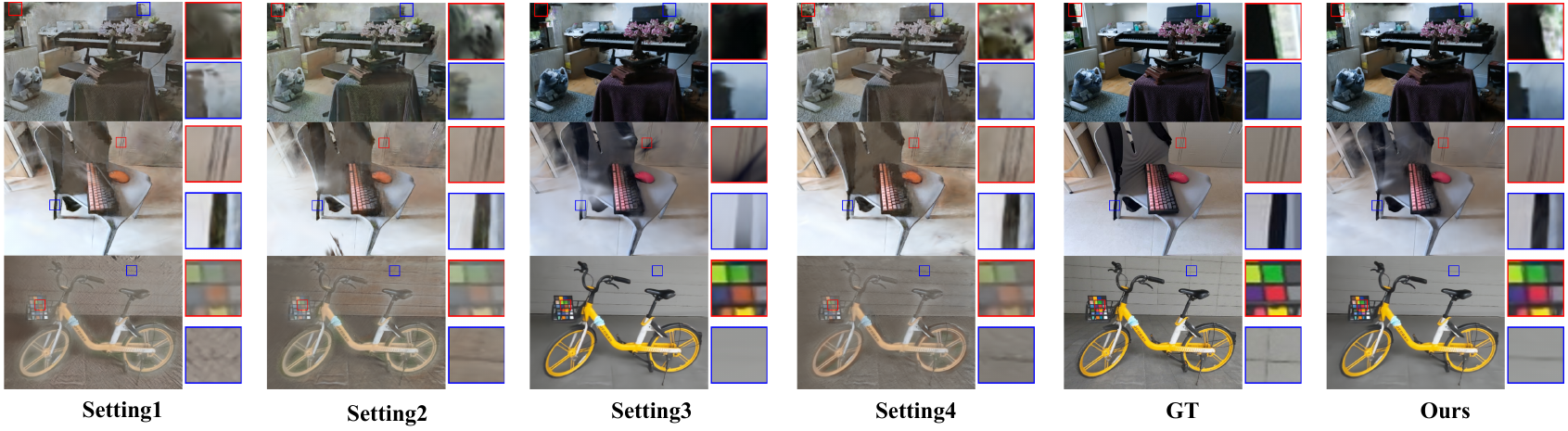}
  \caption{Qualitative comparisons on representative scenes (bike, chair, and bonsai) from the NeRF360, LOM\_full, and LLD datasets, along with comparative analysis against four ablation settings.}
  \label{fig6}
\end{figure}

\noindent\textbf{Break-down Ablation.}
To evaluate the contribution of each component, we conduct a systematic ablation study in \cref{tab:3}. Setting 1 employs only the Baseline and yields relatively low performance. Incorporating ERID (Setting 2) improves the results, indicating that explicit illumination–reflectance separation stabilizes feature representations. Removing IS-FGA (Setting 3) reduces PSNR by approximately \textbf{0.55} dB compared to the full model, with weakened dark-region structure recovery and noise suppression, validating the importance of frequency gating under low-light conditions. Further removing the Reflection Head (Setting 4) decreases PSNR to \textbf{17.56} dB and introduces noticeable cross-scene color and material shifts, demonstrating its critical role in generalization. The complete model (Ours) achieves the best performance on NeRF360, LOM\_full, and LLD. Compared with Setting 1, PSNR improves by \textbf{7.91} dB, SSIM increases by \textbf{0.125}, and LPIPS decreases by \textbf{0.151}, confirming the complementarity of all components. Results in \cref{tab:4} further show that ERID and IS-FGA provide consistent gains during base training. The corresponding qualitative results are shown in \cref{fig6}.

\noindent\textbf{Reflection Head Data Ablation.}
\begin{table*}[t]
\centering
\caption{Comparison of average runtime and performance under different few-shot input scales and optimization iterations.}
\label{tab:5}
\footnotesize
\setlength{\tabcolsep}{3pt}
\renewcommand{\arraystretch}{1.15}
\resizebox{\textwidth}{!}{%
\begin{tabular}{
  l c
  S[table-format=2.2] S[table-format=1.3] S[table-format=1.3] c|
  S[table-format=2.2] S[table-format=1.3] S[table-format=1.3] c|
  S[table-format=2.2] S[table-format=1.3] S[table-format=1.3] c
}
\toprule
\multirow{2}{*}{\textbf{Method}} &
\multirow{2}{*}{\textbf{iters}} &
\multicolumn{4}{c|}{\textbf{NeRF360}} &
\multicolumn{4}{c|}{\textbf{LLD}} &
\multicolumn{4}{c}{\textbf{Average}} \\
\cmidrule(lr){3-6}\cmidrule(lr){7-10}\cmidrule(lr){11-14}
& &
\textbf{PSNR} & \textbf{SSIM} & \textbf{LPIPS} & \textbf{TIME} &
\textbf{PSNR} & \textbf{SSIM} & \textbf{LPIPS} & \textbf{TIME} &
\textbf{PSNR} & \textbf{SSIM} & \textbf{LPIPS} & \textbf{TIME} \\
\midrule

\multirow{2}{*}{\textit{30-imgs}} & 4000 &
24.13 & 0.759 & 0.139 & $\sim$22\,min &
21.75 & 0.752 & 0.180 & $\sim$21\,min &
22.94 & 0.756 & 0.160 & $\sim$21\,min \\
& 800 &
24.07 & 0.756 & 0.145 & \first{$\sim$5\,min} &
21.74 & 0.751 & 0.187 & \first{$\sim$5\,min} &
22.91 & 0.754 & 0.166 & \first{$\sim$5\,min} \\

\multirow{2}{*}{\textit{20-imgs}} & 2000 &
23.83 & 0.749 & 0.150 & $\sim$12\,min &
21.26 & 0.735 & 0.203 & $\sim$12\,min &
22.55 & 0.742 & 0.177 & $\sim$12\,min \\
& 800 &
24.18 & 0.757 & 0.143 & \first{$\sim$5\,min} &
21.72 & 0.749 & 0.189 & \first{$\sim$5\,min} &
22.95 & 0.753 & 0.166 & \first{$\sim$5\,min} \\

\textbf{Ours} & 800 &
23.80 & 0.746 & 0.154 & \first{$\sim$5\,min} &
21.43 & 0.743 & 0.191 & \first{$\sim$5\,min} &
22.62 & 0.745 & 0.173 & \first{$\sim$5\,min} \\

\bottomrule
\end{tabular}
}
\end{table*}
\cref{tab:5} analyzes the impact of different few-shot adaptation configurations on NeRF360 and LLD. The default setting (10 images + 800 iterations) is compared with combinations of 20 or 30 images and 800, 2000, or 4000 optimization iterations. The results indicate that 10 images with 800 iterations achieve the best trade-off among synthesis quality, computational cost, and the number of adaptation images. Although increasing the number of adaptation images or extending the optimization brings marginal improvements, the gains are limited while computational overhead increases significantly. The default configuration maintains a lightweight adaptation setup and enables stable and efficient novel view synthesis with few-shot adaptation.

\section{Conclusion}
We propose MERID-GS, a framework for low-light novel view synthesis. It integrates Multi-Scale Explicit Retinex Illumination Decoupling with 3D Gaussian Splatting to mitigate geometric and photometric inconsistencies. Based on Retinex theory, we explicitly separate illumination and reflectance in the image domain and introduce a learnable gain for adaptive illumination compensation. An Illumination-State-Guided Frequency Gated Attention (IS-FGA) suppresses noise propagation and enhances dark-region structures. Combined with a lightweight Reflection Head, the framework enables fast and stable cross-scene synthesis under sparse views. Experiments on multiple datasets show that MERID-GS outperforms existing methods in synthesis quality and cross-scene generalization, especially in few-shot settings. The LLD dataset provides a new benchmark for low-light novel view synthesis. Future work will explore more extreme low-light conditions and larger-scale scenes.

%
%
\bibliographystyle{splncs04}
\bibliography{main}
\end{document}